\documentclass[11pt, a4paper, logo, internal]{googleresearch}
\pdftrailerid{redacted}
\makeatletter
\renewcommand\bibentry[1]{\nocite{#1}{\frenchspacing\@nameuse{BR@r@#1\@extra@b@citeb}}}
\makeatother

\usepackage{kantlipsum, lipsum}
\usepackage{dsfont}
\usepackage{todonotes}

\definecolor{ourred}{HTML}{E13342}
\definecolor{ourblue}{HTML}{6495ed}
\definecolor{Gray}{gray}{0.92}
\definecolor{racing-green}{rgb}{0.0, 0.8, 0.6}
\definecolor{awesome-red}{rgb}{1.0, 0.13, 0.32}
\definecolor{lgrey}{HTML}{787878}
\definecolor{mgrey}{HTML}{656565}
\definecolor{hgrey}{HTML}{9B9B9B}
\definecolor{black}{HTML}{000000}

\usepackage{xcolor, colortbl}

\usepackage{microtype}
\usepackage{amsmath}
\usepackage{amsfonts} 
\usepackage{algorithm}
\usepackage{algorithmic}
\usepackage{graphicx}
\usepackage{subfigure}
\usepackage{tabularray}
\usepackage{wrapfig}
\usepackage{float}
\usepackage{caption}

\usepackage{pifont}
\definecolor{racing-green}{rgb}{0.0, 0.8, 0.6}
\definecolor{awesome-red}{rgb}{1.0, 0.13, 0.32}

\newcommand{\bluecheck}{{\color{racing-green}{\pmb{\checkmark}}}}
\newcommand{\xmark}{\color{awesome-red}\ding{55}}%

\usepackage{newunicodechar}
\usepackage{enumitem}

\usepackage[authoryear, sort&compress, round]{natbib}

\usepackage[utf8]{inputenc} %
\usepackage[T1]{fontenc}    %
\usepackage{hyperref}       %
\hypersetup{
  colorlinks,
  citecolor=blue,
  linkcolor=red,
  urlcolor=blue}
\usepackage{url}            %
\usepackage{booktabs}       %
\usepackage{amsfonts}       %
\usepackage{nicefrac}       %
\usepackage{microtype}      %
\usepackage{wrapfig} %
\usepackage{graphicx} %
\usepackage{soul} %
\usepackage{enumitem}
\usepackage{amssymb,amsmath}
\usepackage{tikz, adjustbox}
\usepackage[most]{tcolorbox}
\usepackage{subcaption}
\usepackage{booktabs}
\usepackage{multirow}
\usepackage{arydshln}
\usepackage{float}
\usepackage{algorithm}
\usepackage{algorithmic}
\usepackage[capitalize,noabbrev]{cleveref}
\usepackage{fancyvrb}

\DefineVerbatimEnvironment{Code}{BVerbatim}{baseline=t}

\definecolor{codegreen}{rgb}{0,0.6,0}
\definecolor{codegray}{rgb}{0.5,0.5,0.5}
\definecolor{codepurple}{rgb}{0.58,0,0.82}
\definecolor{backcolour}{rgb}{0.95,0.95,0.92}
\lstdefinestyle{mystyle}{
    backgroundcolor=\color{backcolour},   
    commentstyle=\color{codegreen},
    keywordstyle=\color{magenta},
    numberstyle=\tiny\color{codegray},
    stringstyle=\color{codepurple},
    basicstyle=\ttfamily\scriptsize,
    breakatwhitespace=false,         
    breaklines=true,                 
    captionpos=b,                    
    keepspaces=true,                 
    numbers=left,                    
    numbersep=5pt,                  
    showspaces=false,                
    showstringspaces=false,
    showtabs=false,                  
    tabsize=2,
    frame=none,
    aboveskip=1pt,
    belowskip=1pt,
}
\lstdefinestyle{plainins}{
    backgroundcolor=\color{white},   
    commentstyle=\color{codegreen},
    keywordstyle=\color{magenta},
    numberstyle=\tiny\color{codegray},
    stringstyle=\color{codepurple},
    basicstyle=\ttfamily\scriptsize,
    breakatwhitespace=false,         
    breaklines=true,                 
    captionpos=b,                    
    keepspaces=true,                 
    numbers=none,                    
    numbersep=5pt,                  
    showspaces=false,                
    showstringspaces=false,
    showtabs=false,                  
    tabsize=2,
    aboveskip=0pt,
    belowskip=0pt,
    frame=single
}
\lstdefinestyle{plainexam}{
    backgroundcolor=\color[HTML]{FFFCF3},   
    commentstyle=\color{codegreen},
    keywordstyle=\color{magenta},
    numberstyle=\tiny\color{codegray},
    stringstyle=\color{codepurple},
    basicstyle=\ttfamily\scriptsize,
    breakatwhitespace=false,         
    breaklines=true,                 
    captionpos=b,                    
    keepspaces=true,                 
    numbers=none,                    
    numbersep=5pt,                  
    showspaces=false,                
    showstringspaces=false,
    showtabs=false,                  
    tabsize=2,
    aboveskip=0pt,
    belowskip=0pt
}

\title{Batch Calibration: Rethinking Calibration for In-Context Learning and Prompt Engineering
}

\correspondingauthor{ \{hzhouml, subhrajitroy\}@google.com. Work done while Han Zhou was a Student Researcher at Google Research. }

\renewcommand{\today}{}

\author[1,3]{Han Zhou}
\author{Xingchen Wan}
\author[2]{Lev Proleev}
\author[2]{Diana Mincu}
\author[2]{Jilin Chen}
\author[1]{Katherine A Heller}
\author[2]{\\Subhrajit Roy}

\affil[1]{Google Research}
\affil[2]{Google DeepMind}
\affil[3]{University of Cambridge}

\begin{abstract}
Prompting and in-context learning (ICL) have become efficient learning paradigms for large language models (LLMs). However, LLMs suffer from prompt brittleness and various bias factors in the prompt, including but not limited to the formatting, the choice verbalizers, and the ICL examples. To address this problem that results in unexpected performance degradation, calibration methods have been developed to mitigate the effects of these biases while recovering LLM performance. In this work, we first conduct a systematic analysis of the existing calibration methods, where we both provide a unified view and reveal the failure cases. Inspired by these analyses, we propose \emph{Batch Calibration} (BC), a simple yet intuitive method that controls the contextual bias from the batched input, unifies various prior approaches, and effectively addresses the aforementioned issues. BC is zero-shot, inference-only, and incurs negligible additional costs. In the few-shot setup, we further extend BC to allow it to \emph{learn} the contextual bias from labeled data. We validate the effectiveness of BC with PaLM 2-(S, M, L) and CLIP models and demonstrate state-of-the-art performance over previous calibration baselines across more than 10 natural language understanding and image classification tasks.
\end{abstract}

\begin{document}

\maketitle

\section{Introduction}
\label{intro}

\begin{wrapfigure}{r}{0.48\textwidth}
  \begin{center}
  \vspace{-1.2cm}\includegraphics[width=0.48\textwidth]{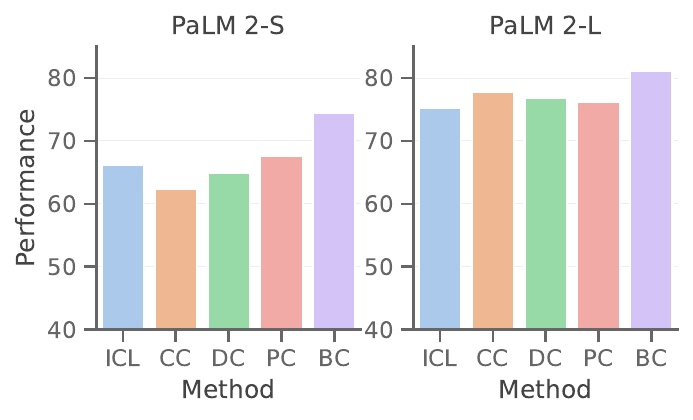}
  \end{center}
  \vspace{-0.58cm}
  \caption{Batch Calibration (BC) achieves the best performance on 1-shot ICL over calibration baselines on an average of 13 classification tasks on PaLM 2-S and PaLM 2-L \citep{anil2023palm}.}
  \vspace{-0.6cm}
\end{wrapfigure}
Prompting large language models (LLMs) \citep{chowdhery2022palm, anil2023palm} has become an efficient learning paradigm for adapting LLMs to a new task by conditioning on human-designed instructions. The remarkable in-context learning (ICL) ability of LLMs also leads to efficient few-shot learners that can generalize from few-shot input-label pairs \citep{brown2020language, liu2023pre}. However, the predictions of LLMs are highly sensitive and even biased to the choice of templates \citep{min-etal-2022-rethinking}, verbalizers \citep{holtzman-etal-2021-surface}, and demonstrations \citep{liu-etal-2022-makes}, resulting in barriers for pursuing efficiently adaptable and robust LLM applications. 

Extensive research has been devoted to mitigating these biases, which we explicitly refer to the a-priori propensity of LLMs to predict certain classes over others unfairly. \citet{lu-etal-2022-fantastically} provide an analysis of the impacts of the order of ICL examples to LLMs and have explored the order selection mechanisms for ICL. On the other hand, \citet{zhao2021calibrate} reveal the bias of language models toward certain answers and propose to calibrate the LLM given content-free tokens. More recently, \citet{fei-etal-2023-mitigating} detect the domain-label bias, and \citet{han2023prototypical} treat the calibration of LLMs, a technique for mitigating the label bias, as learning a robust decision boundary. Though multiple calibration solutions have been provided, the field currently lacks a unified analysis that systematically distinguishes and explains the unique characteristics, merits, and downsides of each approach.

In this work, we first conduct a comprehensive analysis across existing calibration methods for LLMs. We approach the calibration problem from a distinctive point of view by interpreting the decision boundaries for each calibration method together with the ICL decision boundary. We start observing fatal failure cases for each method by extending them to more challenging and under-explored evaluation tasks. We then conclude the current limitation for each method with a novel interpretation from the decision boundary perspective, pointing to the need for a unified and widely applicable solution for conquering diverse bias sources in the field of LLM efficient learning. 

Inspired by these findings, we propose \emph{Batch Calibration} (BC), a zero-shot and inference-only calibration method for prompting and ICL. The central objective of BC is to accurately model the bias from the prompt context (referred to as \textit{contextual bias} in this paper) by marginalizing the LLM scores in the batched input. The simplicity of the design of BC only brings negligible computation overhead at the output of the LLM. We further extend BC to the black-box few-shot learning (BCL), a practical case where labeled data is available, by introducing a \emph{single} learnable parameter into BC, which enables it to adapt and \emph{learn} the contextual bias from the available data resources.

We conducted extensive experiments on more than 10 natural language understanding tasks together with image classification tasks. BC stands as the most widely applicable calibration method while achieving state-of-the-art results. With the proposed black-box few-shot BCL framework, we show that further slight gains can be achieved by leveraging more labeled data. We provide further analysis with BC on robustness with templates, ICL choices and orders, and verbalizers, validating that BC can effectively alleviate prompt brittleness and make prompt engineering easier. To summarize, we provide the following contributions:
\begin{itemize}[leftmargin=*]
    \item We provide a unified and systematic analysis of existing calibration methods through their decision boundaries, investigate the common use of content-free tokens as an estimator of contextual bias, and identify their deficiency with individual case studies. 
    \item We propose Batch Calibration (BC), a zero-shot and inference-only calibration method for ICL, that mitigates the bias from the batch. We further extend BC to learn from few-shot data.
    \item We show that while conceptually simple, BC attains state-of-the-art performance in both zero-shot and few-shot learning setups over widely selected tasks with PaLM-2 and CLIP models.
\end{itemize}

\section{Related Work}
\label{sec:related_work}

\paragraph{Understanding and Improving ICL.} \citet{lu-etal-2022-fantastically} show the sensitivity of LLMs to ICL examples. This phenomenon is further explained through the effect from pretraining term frequencies \citep{razeghi-etal-2022-impact} and corpora \citep{shin-etal-2022-effect}. Meanwhile, \citet{xie2022an} explain the ICL process through implicit Bayesian inference, and \citet{wei2023larger} show the emergent ability of LLMs by learning new input-label mappings. Various methods have been proposed to optimally select better in-context templates \citep{sorensen-etal-2022-information, pan-etal-2023-context, yin-etal-2023-read} and examples \citep{rubin-etal-2022-learning, liu-etal-2022-makes, wan2023universal}. Specifically, \citet{wan-etal-2023-better} introduce a selection criteria for in-context examples based on the confidence. Recently, noisy channel prompting \citep{min-etal-2022-noisy}, flipped learning \citep{ye2022guess}, and self-ensembling \citep{li2023task} have been proposed for robust ICL. Learning to assign labels by k-nearest neighbors \citep{xu2023knn} and training decoder networks \citep{cui-etal-2023-decoder} are also effective alternatives for few-shot ICL. 

\paragraph{Bias in ICL and Calibrating LLMs.}
\citet{zhao2021calibrate} reveal the instability of LLMs in few-shot learning and demonstrate three bias sources: majority label bias, recency bias, and common token bias, as the bias factors behind the instability. They propose contextual calibration (CC) to mitigate these biases by grounding the prediction based on a content-free token as sample inputs.
\citet{si-etal-2023-measuring} characterize the feature bias of LLMs, and \citet{wang2023large} introduce the positional bias in candidate choices. \citet{fei-etal-2023-mitigating} further observe the existence of domain-label bias and propose domain-context calibration (DC) that uses random in-domain tokens for estimating the bias. Meanwhile, \citet{han2023prototypical} analyze the impact of decision boundary for text classification tasks and propose to estimate prototypical clusters by Gaussian mixture models, thereby learning a robust decision boundary. Concurrently with our work, \citet{pezeshkpour2023large} spot the positional bias in multiple-choice questions, and \citet{zheng2023large} propose to debias the positional bias in multiple choices with permutation-based prior estimation. We will discuss quantitatively and provide a unified analysis of these methods in Sec.~\ref{Pre}. As we will show, our proposed method differentiates from these methods as a generalizable solution across challenging classification tasks and modalities.

\section{A Systematic Analysis of Calibration}
\label{Pre}
\subsection{Bias in Prompting and In-Context Learning (ICL)}
\label{sec:notation}
Prompting is an efficient learning paradigm that allows LLMs to perform zero-shot inference by conditioning on a human-designed instruction. Formally, denoting a test query-target pair $\{x_i, y_i\}$ and instruction as the context $C$ for a classification task, LLMs make prediction by computing: $\operatorname*{arg\,max}_{y\in \mathcal{Y}}\mathbf{p}(y|x_i, C)$, where $\mathbf{p} \in \mathbb{R}^J$ is the logits, and $\mathcal{Y}$ denotes the verbalizers that define the label set for $J$ classes. ICL further enables LLM to learn from $k$ input-label pairs (i.e., few-shot setup), $s^{(i)} = \texttt{Template}(x^{(i)}, y^{(i)}) \, \forall i \in\{1,...,k\}$, by concatenating few-shot demonstrations in a pre-defined template as the context, $C= \texttt{Concat}(s^{(i)},...,s^{(k)})$. Though ICL has demonstrated strong performance with easy implementations, the prediction of LLMs is shown to be biased towards certain answers due to different elements of $\mathbf{p}(y|x_i, C)$ \citep{lu-etal-2022-fantastically}. In the ICL context $C$, majority label bias and recency label bias \citep{zhao2021calibrate} can bias the prediction of LLMs toward the most frequent label and the label towards the end of the demonstration, respectively. Among verbalizer tokens $y_j\in \mathcal{Y}$, LLMs are shown to be inherently biased towards predicting the label-tokens that appear more frequently from pretraining term statistics \citep{shin-etal-2022-effect, razeghi-etal-2022-impact}. These bias factors significantly degrade the performance of LLMs for robust ICL applications.

\subsection{Overview of ICL Calibration Methods.}
As introduced in Sec.~\ref{sec:related_work}, various \emph{calibration} methods have been proposed to mitigate the issue of bias identified above. In this section, we provide an overview of the state-of-the-art calibration methods. 
\paragraph{Contextual Calibration \citep{zhao2021calibrate} (CC).}
Motivated by a common calibration technique that applies affine transformation on the model outputs \citep{platt1999probabilistic, guo2017calibration}, \citet{zhao2021calibrate} propose to calibrate the LLM prediction by first measuring the entire test-time distribution $\hat{\mathbf{p}}$ by a content-free input. Using ``\texttt{N/A}'' as a content-free example, the model score distribution is generated by $\hat{\mathbf{p}}_{\text{cf}} := \mathbf{p}(y|[\texttt{N/A}], C)$. CC then generates the calibrated output by transforming the uncalibrated scores $\mathbf{p}(y|x, C)$ with $\mathbf{W} \in \mathbb{R}^{J \times J}$ via $\mathbf{W}\mathbf{p}(y|x, C)$, where $\mathbf{W}=\operatorname*{diag}(\hat{\mathbf{p}}_{\text{cf}})^{-1}$ offsets the uncalibrated scores with the model score (a contextual prior) triggered by the content-free sample.

\paragraph{Domain-Context Calibration \citep{fei-etal-2023-mitigating} (DC).}
Instead of using a single content-free token, \citet{fei-etal-2023-mitigating} propose DC that estimates a contextual prior $\hat{\mathbf{p}}(y|C)$ by using a random in-domain sequence. It randomly sampled $L$ tokens at an average sentence length from an unlabeled text set. Then, it estimates the content-free prediction prior by averaging the model score $T$ times, such that: $\hat{\mathbf{p}}_{\text{random}}=\frac{1}{T}\sum_{t=1}^{T} \mathbf{p}(y|[\textsc{Random Text}]_{t}, C)$. The final test-time prediction is then calibrated by dividing the estimated prior prediction, or equivalently in logits space, $\mathbf{p}(y|x_i, C)-\hat{\mathbf{p}}_{\text{random}}$. 

\paragraph{Prototypical Calibration \citep{han2023prototypical} (PC).}
PC learns a decision boundary with Gaussian mixture models (GMMs). It estimates $J$ prototypical clusters for the model output $\mathbf{p}$ for $J$ classes: $P_{\text{GMM}}(\mathbf{p})  =\sum_{j=0}^{J-1}\alpha_{j} P_{G}(\mathbf{p}|\mathbf{\boldsymbol\mu_{j}},\mathbf{\boldsymbol\Sigma_{j}})$, where $P_{G}$ denotes a multi-variate Gaussian distribution, and the parameters: mixing coefficient $\alpha$, mean vector $\mathbf{\boldsymbol\mu}$, covariance matrix $\mathbf{\boldsymbol\Sigma}$ are estimated by the Expectation-Maximization \citep{moon1996expectation}. Followed by an automatic label assignment strategy, the predicted label is then computed by $\operatorname*{arg\,max}_{j}P_{G}(\mathbf{p}_j|\mathbf{\mu^{*}},\mathbf{\boldsymbol\Sigma^{*}})$ in the inference time. This EM-GMM process can require up to $T$ repetitions to stabilize its estimation of clusters where $T$ is a hyperparameter of the algorithm. 
\begin{table}
\caption{Calibration methods with their mathematical formulation and their equivalent decision boundary derivations in a two-dimensional problem. The cost for the number of API calls is denoted as \#Forward, where $1$ counts for the original ICL forward cost, and PC and BC incur no additional API costs. We refer derivations of decision boundaries to Sec.~\ref{sec:analysis}. The potential failure case for each calibration method in practical scenarios is marked as \xmark\color{black}.}
\label{tab:analysis}
\centering
\resizebox{\linewidth}{!}{%
\begin{tblr}{
  column{2-9} = {c},
  vline{2} = {-}{},
  hline{1-2,6} = {-}{},
  hline{1,6} = {-}{0.08em},
  stretch = 0.5
}
\shortstack{Method\\ \hfill}    & \shortstack{Token\\ \hfill} &\shortstack{\#Forward\\ \hfill} & \shortstack{Comp.\\Cost}   & \shortstack{Cali.\\ Form} & \shortstack{Learning\\ Term} & \shortstack{Decision\\ Boundary $h(\mathbf{p})$}&  \shortstack{Multi-\\Sentence} & \shortstack{Multi-\\Class} \\
CC        & N/A &$1+1$         & Inverse               & $\mathbf{W}\mathbf{p}+\mathbf{b}$       &  $\mathbf{W}=\operatorname*{diag}(\hat{\mathbf{p}})^{-1}$, $\mathbf{b}=\mathbf{0}$      &$p_0=\alpha p_1$              & \xmark           &  \bluecheck           \\
DC        & Random &$20+1$        & Add               & $\mathbf{W}\mathbf{p}+\mathbf{b}$         & $\mathbf{W}=\mathbf{I}$, $\mathbf{b}=-\frac{1}{T}\sum_{t} \mathbf{p}(y|\text{text}_{j}, C)$ & $p_0=p_1+\alpha$      &   \xmark         &   \bluecheck          \\
PC        & - &$1$         & EM-GMM &      -       &  $\sum_{j}\alpha_{j} P_{G}(\mathbf{p}|\mathbf{\boldsymbol\mu_{j}},\mathbf{\boldsymbol\Sigma_{j}})$   &$P_\text{G}(\mathbf{p}|\mu_{0},\Sigma_{0})=P_\text{G}(\mathbf{p}|\mu_{1},\Sigma_{1})$      &    \bluecheck        &    \xmark         \\
BC (Ours) & - &$1$         &  Add              & $\mathbf{W}\mathbf{p}+\mathbf{b}$   & $\mathbf{W}=\mathbf{I}$, $\mathbf{b}=-\operatorname*{\mathbb{E}}_{
x}\Big[\mathbf{p}(y|x, C) \Big]$ &$p_0=p_1+\alpha$    &  \bluecheck          &  \bluecheck           
\end{tblr}}
\end{table}

\begin{figure*}[t!]
  \vspace{-0.4cm}
  \begin{subfigure}{}
    \centering\includegraphics[width=\linewidth]{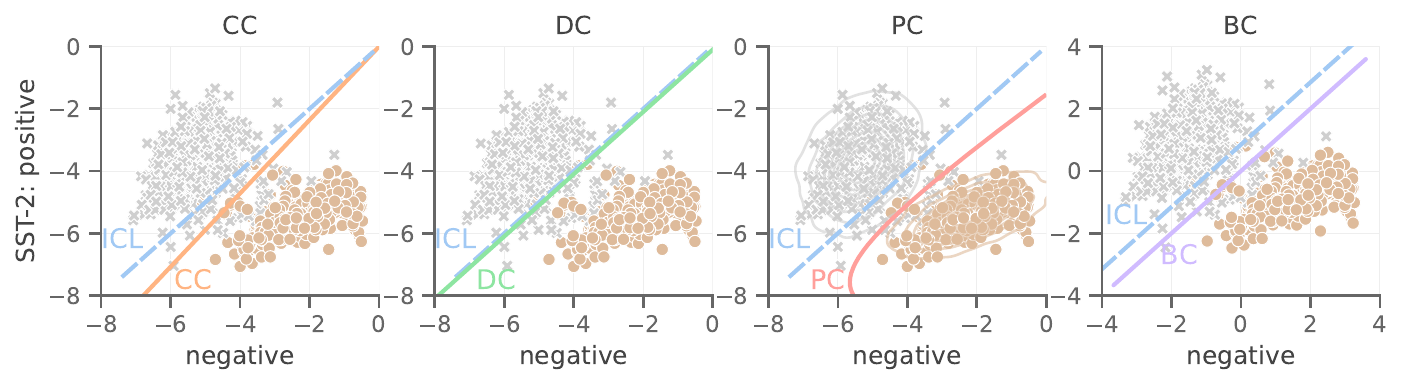}
    \caption*{}
  \end{subfigure}
  \vspace{-1.3cm}
  \begin{subfigure}{}
    \centering\includegraphics[width=\linewidth]{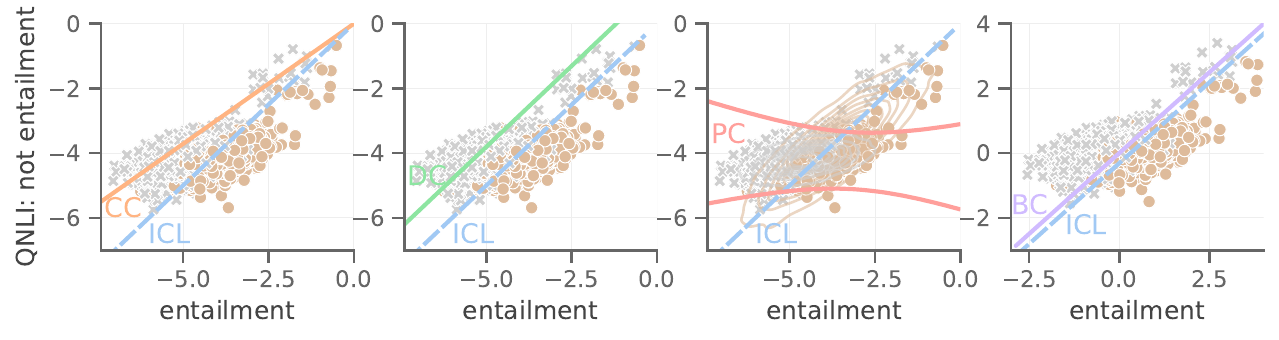}
    \caption*{}
  \end{subfigure}
  \vspace{-1cm}
  \caption{Visualization of the decision boundaries of uncalibrated ICL, and after applying existing calibration methods and the proposed BC (to be introduced in Sec~\ref{fig:method}) in representative binary classification tasks of SST-2 (top row) \citep{socher-etal-2013-recursive} and QNLI (bottom row) \citep{wang-etal-2018-glue} on 1-shot PaLM 2-S. We show success and failure cases for each baseline method (CC, DC, and PC), whereas BC is consistently effective. Refer to Appendix \S\ref{app:aexp} for more examples.}
  \label{fig:db}
\end{figure*}
\subsection{Design Principles Behind Calibrations}
\label{sec:analysis}

Summarizing the calibration methods with distinctive design principles discussed so far, in Table~\ref{tab:analysis}, we present a unified view of the characteristics of each method with their mathematical formulation, computation cost, and strengths \& weaknesses. Though each approach demonstrates a clear motivation for calibrating ICL, it is still unclear which method surpasses others in what scenarios. We proceed with an in-depth analysis of existing methods in representative tasks. We provide a novel view of calibration methods from a multi-variate decision boundary perspective. In pursuit of practical guidelines for ICL calibration, we set out two important research questions behind their design principles: \textbf{1)} What constitutes a better decision boundary for calibrations? \textbf{2)} Is content-free prior a good estimator of contextual bias?

\paragraph{What Constitutes a Better Decision Boundary for Calibrations?}
To address this research question, we first derive the decision boundary for each category of calibration methods. We recall that the classification by a LLM is based on $\operatorname*{arg\,max}_{j \in \{0,...,J-1\}}p_j$ where $p_j$ denotes the $j$-th element of output vector $\mathbf{p}$. Consider binary classification problem for simplicity: the decision boundary $h(\mathbf{p})$ for ICL is given by the line $p_0 - p_1 = 0$: the model predicts class 0, $y_0$, if $p_0 - p_1 \geq 0$, and class 1 otherwise. Consequently, CC and DC that apply an affine transformation at $\mathbf{p}$ is equivalent to a linear transformation to the decision boundary. In CC with $\mathbf{W}=\operatorname*{diag}(\hat{\mathbf{p}})^{-1}$, $\mathbf{b}=\mathbf{0}$, the decision boundary can then be derived as $
    p_0 \times \frac{1}{\hat{p}_{0}} = p_1 \times \frac{1}{\hat{p}_{1}} \rightarrow p_0 - p_1 \times \frac{\hat{p}_{0}}{\hat{p}_{1}} = 0$,
which is a \textit{rotation} of the ICL's linear decision boundary around the origin. Similarly, DC with $\mathbf{W}=\mathbf{I}$, $\mathbf{b}=-\frac{1}{T}\sum_{t} \mathbf{p}(y|[\textsc{Random Text}]_{t}, C)=-\hat{\mathbf{p}}$ is equivalent to a \textit{shift} of ICL's linear decision boundary away from the origin, such that $p_0-p_1 =  (\hat{p}_{0}-\hat{p}_{1})$. It is worth noting that both calibration choices lead to a linear decision boundary, indicating that the calibration problem can be framed as an unsupervised decision boundary learning problem. Based on this intuition, we further derive the decision boundary for PC as $P_\text{G}(\mathbf{p}|\mu_{0},\Sigma_{0})-P_\text{G}(\mathbf{p}|\mu_{1},\Sigma_{1})=0$, which delivers a non-linear boundary between the estimated Gaussian mixtures. We conduct a preliminary experiment to visualize the derived decision bounds from existing calibration methods alongside the ICL baseline. In Fig.~\ref{fig:db}, we observe that uncalibrated ICL is biased towards predicting \texttt{negative} in the SST-2 task. This biased prediction is then mitigated by each calibration method, where we observe a rotated decision boundary from CC, a shifted boundary from DC, and a non-linear boundary between the GMMs by PC. However, in the QNLI task (bottom row of Fig.~\ref{fig:db}), we observe failure cases in the calibration baselines, in particular, PC (third figure from the left), where it fails to capture the correct distribution for each class. From Fig.~\ref{fig:db} and the additional results in Fig.~\ref{fig:dba} in Appendix~\S\ref{app:aexp}, we find that while theoretically more flexible, the non-linear decision boundaries learned by PC tend to be susceptible to overfitting and may suffer from instability in EM-GMM. \begin{wrapfigure}{r}{0.48\textwidth}
  \begin{center}
  \vspace{-0.75cm}
  \includegraphics[width=0.48\textwidth]{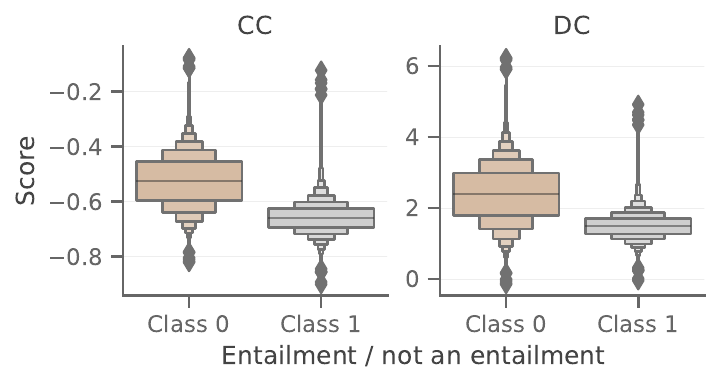}
  \end{center}
  \vspace{-0.6cm}
  \caption{The distribution of ICL scores after applying CC and DC on QNLI. Due to an unfair content-free prior, the prediction by 1-shot PaLM-2 is biased towards \texttt{entailment}.}
  \label{fig.case}
    \vspace{-1.5cm}
\end{wrapfigure} We hypothesize that the PC boundary is even more vulnerable to instability for more challenging multi-class tasks due to the increased difficulties of learning clusters and assigning classes correctly. Conversely, we find that linear decision boundaries, as evidenced by CC and DC, can be more robust and generalizable across tasks. We validate this hypothesis by proposing BC with extensive experiments in Sec.~\ref{sec:mainexp}.

\paragraph{Is Content-free Input a Good Estimator of the Contextual Prior?}

CC and DC both use a linear decision boundary but differ from each other by leveraging different formats of a content-free input to estimate the contextual prior. However, as we observed in Fig.~\ref{fig:db}, they both exhibit failure cases in 
QNLI, a question-answering NLI task. We hypothesize that contrary to the proposals made by CC and DC, relying on content-free tokens for calibration is \textit{not} always optimal and may even introduce additional bias, depending on the task type.
For example, in a textual entailment task involving question-sentence pairs, we empirically observe that an ICL template employed with a content-free token `\texttt{N/A}' such as `\texttt{Question: N/A, Sentence: N/A, Answer:}' will result in a biased prediction towards `\texttt{entailment}', because although `\texttt{N/A}' is intended to be content-free, the LLM may nevertheless construe `\texttt{N/A}' in the sentence to be substantively entailed to the `\texttt{N/A}' in the question due to surface text equivalency.
This phenomenon holds true for other multi-text classification tasks, such as paraphrasing and word disambiguation tasks. Consequently, the prior estimated via a single content-free token can lead to further bias. DC introduces multiple randomly sampled tokens to form a content-free input, e.g. `\texttt{Question: that What old rubisco's the did Which?}'. We suspect a possible reason is that random sequences, when used in conjunction with in-context demonstrations, can be susceptible to spurious relations among them that often lead to unfair priors further skewing the predictions, which is also reflected in Fig.~\ref{fig.case}, where CC and DC fail to mitigate the contextual bias in the QNLI task. In sum, the empirical observation shows that content-free inputs can be inappropriate prior estimators, especially for multi-sentence classification tasks.

\section{Batch Calibration}
\label{sec:method}
\begin{figure*}[!t]
    \centering
    \includegraphics[width=\linewidth]{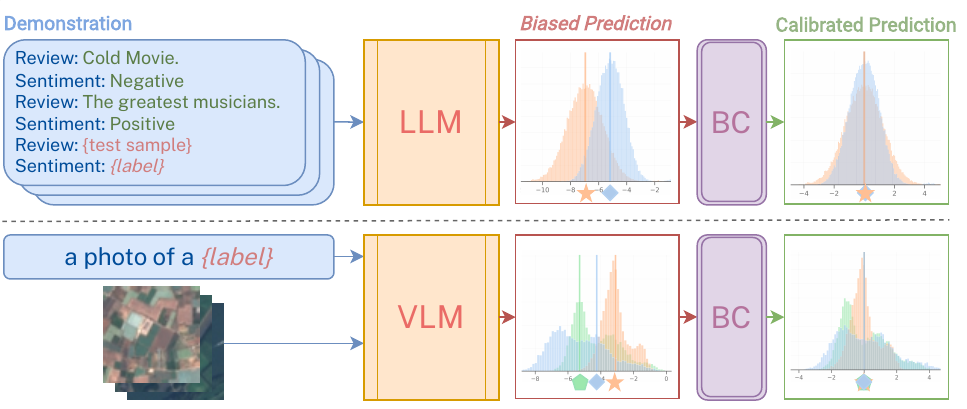}
          \vspace{-0.5cm}
    \caption{Illustration of Batch Calibration (BC). Batches of demonstrations with in-context examples and test samples are passed into the LLM. Due to implicit bias sources in the context, the score distribution from the LLM becomes highly biased. BC is a modular and adaptable layer option appended to the output of the LLM/VLM. BC generates calibrated scores according to Eq. \ref{eq:mean} \& \ref{eq:overprior}. Highlighted symbols indicate the distribution means (visualized \textit{for illustration only}). }
    \label{fig:method}
\end{figure*}

Inspired by the previous discussions, we now propose Batch Calibration (BC), a zero-shot, inference-only, and generalizable calibration technique with negligible computation cost. We further discuss how BC can be extended to \textit{vision}-language models as well as the black-box \textit{few-shot} learning setup.
\paragraph{Batch Calibration (BC).}
Following the discussion in Sec.~\ref{sec:analysis}, we argue that the most critical component for calibration is to accurately estimate the contextual bias term $\mathbf{p}(y|C)$. Both CC and DC, which use content-free and in-domain random tokens as trigger signals, respectively, have failure cases in multi-sentence classification when the estimation of the contextual bias is inaccurate. On the other hand, PC is vulnerable to overfitting and may incorrectly model the mixtures, especially in high-dimensional space. We, therefore, opt for a linear decision boundary for its robustness, and instead of relying on content-free tokens, we propose to estimate the contextual bias for each class $\mathbf{p}(y=y_j|C)$ from a batch with $M$ samples, $\{x^1, ..., x^M\}$, in a \textit{content-based} manner by marginalizing the output score over all samples $x\sim P(x)$ within the batch:
\begin{align}
    \mathbf{p}(y|C)_{j}  =\operatorname*{\mathbb{E}}_{
x\sim P(x)}\Big[\mathbf{p}(y=y_j|x, C) \Big] \approx \frac{1}{M}\sum_{i=1}^{M} \mathbf{p}(y=y_j|x^{(i)}, C) \, \forall \, y_j \in \mathcal{Y}.
\label{eq:mean}
\end{align}
We then obtain the calibrated probability by dividing the output probability over the contextual prior, which is equivalently by shifting the log-probability by the estimated mean of each class:
\begin{align}
    \hat{y}_i = \operatorname*{arg\,max}_{y\in \mathcal{Y}} \mathbf{p}_{\text{BC}}(y|x_i, C)=\operatorname*{arg\,max}_{y\in \mathcal{Y}}\big[\mathbf{p}(y|x_i, C)-\hat{\mathbf{p}}(y|C)\big].
\label{eq:overprior}
\end{align}
It is noteworthy that this BC procedure is zero-shot and only involves unlabeled test samples. BC incurs negligible computation costs. We may either compute the correction term $\hat{\mathbf{p}}(y|C)$ once all test samples are seen or, alternatively, in an on-the-fly manner that dynamically processes the outputs. To do so, we may use a running estimate of the contextual bias for BC. At the $n+1$ mini-batch, the bias term is given by: 
\begin{align}
    \mathbf{p}_{\text{r}}^{n+1}(y|C)  =\frac{n}{n+1}\mathbf{p}_{\text{r}}^{n}(y|C) + \frac{1}{n+1} \hat{\mathbf{p}}^{n+1}(y|C),
\end{align}
thereby allowing BC's calibration term to be estimated from a small number of mini-batches that is subsequently stabilized when more mini-batches arrive.

\paragraph{Adjustable Batch Calibration Layer (BCL).}\begin{wrapfigure}{r}{0.48\textwidth}
  \begin{center}
     \centering
     \vspace{-0.92cm}
     \includegraphics[width=\linewidth]{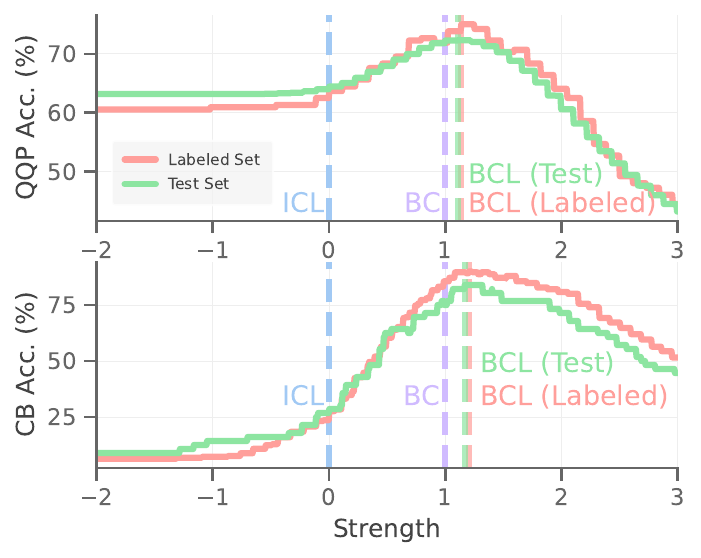}
     \vspace{-0.7cm}
     \caption{\textit{BC benefits from labeled data: }The performance of the adjustable BCL compared to the zero-shot BC with a changing strength. The \texttt{strength} $\gamma$ at 0 and 1 represent the uncalibrated ICL and BC, respectively. We highlight the optimal strength learned from a labeled set and the best test strength. Refer to Appendix \S\ref{app:aexp} for more examples.}\label{fig:bcl}
     \vspace{-1cm}
\end{center}
\end{wrapfigure}
While BC is designed to be zero-shot and inference-only, it is also common that some \textit{labeled} data are available. In this section, we describe a simple, adapted variant of BC that may further refine the calibration and mitigate any estimation errors from the unlabeled data, which we term \textit{BCL}. Specifically, instead of deducting the bias term $\hat{\mathbf{p}}$ from the test data only, 
we introduce a single additional hyperparameter \textit{strength} $\gamma\in \mathbb{R}$:
\begin{align}
    \mathbf{p}_{\text{BCL}}(y|x_i, C)=\mathbf{p}(y|x_i, C)-\gamma\hat{\mathbf{p}}(y|C),
\label{eq:BC+}
\end{align}
where $\gamma$ controls the strength of BC. 
To select the appropriate $\gamma$, we simply perform a grid search by uniformly sampling $T$ different $\gamma$ values in $[a, b]$ (we set $[a, b]:= [-5, 5]$, but any reasonable range may be used). The strength $\gamma$ is then learned by $\gamma^* = \operatorname*{arg\,max}_{\gamma\in[a, b]}R(\mathbf{p}_{\text{BC}
}, \gamma)$, where $R(\cdot,\cdot)$ is the evaluation function (e.g., accuracy) on the set of \textit{labeled} data, allowing the amount of calibration to be adjusted from evaluation metrics directly.

We give concrete examples in Fig. \ref{fig:bcl}, which illustrates the effect of BCL where we plot the accuracy in QQP \citep{wang-etal-2018-glue} and CB \citep{de2019commitmentbank} tasks over a range of $\gamma$. We observe that $\gamma=1$, which corresponds to BC without adjustment (purple line), leads to a strong but not optimal performance. By using the $\gamma$ learned from the labeled data (a 128-shot randomly sampled set in this case), BCL estimates the contextual bias more precisely by leveraging the labeled data and achieves a performance that is very close to the optimal. We refer readers to Table~\ref{tab:bcl} for more results.
\paragraph{Calibrating Vision-Language Models.}
Recently, vision-language models (VLM) \citep{radford2021learning}, which simultaneously encode visual and textual information, have demonstrated strong zero-shot generalization capability by rewriting class labels. However, the sources of bias as LLMs have also been observed in prompting VLMs \citep{alayrac2022flamingo} but have not been adequately addressed. In this work, we propose to apply BC to Zero-Shot (ZS) CLIP \citep{radford2021learning} and mitigate the biases in its zero-shot classifications. We follow the same notation from Sec. \ref{sec:notation}, where the test image is now $x$, and the prompt template becomes the context, $C$. Similarly, we append the BC layer at the output of the ZS CLIP and calibrate for each class following Eq. \ref{eq:mean} \& \ref{eq:overprior}.

\begin{table}
\centering
\caption{Accuracy (\%) on natural language classification tasks with 1-shot PaLM 2-S and PaLM 2-L Models \citep{anil2023palm}. We report the mean and standard deviation for all results for 5 different in-context examples. We reproduce all baselines, and the implementation details are described in Appendix \S \ref{appendix:details}. The \textbf{best} and \textbf{\color{mgrey}{second-best}} results are marked in bold fonts and ranked by color.}
\label{tab:main}
\resizebox{\linewidth}{!}{%
\begin{tblr}{
  width = \linewidth,
  cell{1}{2} = {c=5}{c},
  cell{1}{7} = {c=5}{c},
  cell{2}{2-11} = {c=1},
  vline{2-3} = {1}{},
  vline{2,7} = {1-17}{},
  hline{1-3,16-17} = {-}{},
    hline{1,17} = {-}{0.08em},
  stretch = 0.5
}
Model & PaLM 2-S &  &  &  &  & PaLM 2-L &  &  &  & \\
Method & ICL & CC & DC & PC & BC & ICL & CC & DC & PC & BC \\
SST-2 & \color{mgrey}93.62\textsubscript{0.62} & \color{mgrey}\textbf{95.50}\textsubscript{0.25} & \color{mgrey}94.29\textsubscript{0.32} & \textbf{95.71}\textsubscript{0.10} & \color{mgrey}95.44\textsubscript{0.15} & \color{mgrey}93.16\textsubscript{5.18} & \textbf{95.82}\textsubscript{0.62} & \color{mgrey}94.91\textsubscript{2.01} & \color{mgrey}95.64\textsubscript{0.47} & \color{mgrey}\textbf{95.78}\textsubscript{0.55}\\
MNLI & \color{mgrey}\textbf{68.52}\textsubscript{7.98} & \color{mgrey}60.07\textsubscript{11.26} & \color{mgrey}63.45\textsubscript{1.99} & \color{mgrey}59.29\textsubscript{13.79} & \textbf{75.12}\textsubscript{2.76} & \color{mgrey}72.77\textsubscript{3.65} & \color{mgrey}\textbf{79.45}\textsubscript{3.46} & \color{mgrey}71.53\textsubscript{4.86} & \color{mgrey}78.68\textsubscript{7.10} & \textbf{81.34}\textsubscript{2.29}\\
QNLI & \color{mgrey}\textbf{81.20}\textsubscript{1.90} & \color{mgrey}56.86\textsubscript{3.29} & \color{mgrey}65.62\textsubscript{3.53} & \color{mgrey}69.82\textsubscript{17.73} & \textbf{82.45}\textsubscript{1.82} & \color{mgrey}64.68\textsubscript{3.53} & \color{mgrey}\textbf{69.71}\textsubscript{4.89} & \color{mgrey}68.97\textsubscript{3.27} & \color{mgrey}61.01\textsubscript{15.26} & \textbf{87.90}\textsubscript{1.24}\\
MRPC & \color{mgrey}66.42\textsubscript{10.15} & \color{mgrey}\textbf{70.44}\textsubscript{0.94} & \color{mgrey}68.58\textsubscript{0.21} & \textbf{71.86}\textsubscript{1.29} & \color{mgrey}70.05\textsubscript{2.40} & \color{mgrey}\textbf{73.19}\textsubscript{1.21} & \color{mgrey}72.40\textsubscript{3.53} & \color{mgrey}68.68\textsubscript{0.40} & \textbf{75.39}\textsubscript{2.60} & \color{mgrey}70.39\textsubscript{2.56}\\
QQP & \color{mgrey}63.91\textsubscript{0.66} & \color{mgrey}\textbf{65.55}\textsubscript{5.34} & \color{mgrey}53.92\textsubscript{9.35} & \color{mgrey}65.28\textsubscript{3.42} & \textbf{71.48}\textsubscript{1.46} &\textbf{82.57}\textsubscript{0.75}  & \color{mgrey}81.17\textsubscript{2.03} & \color{mgrey}78.32\textsubscript{1.82} & \color{mgrey}\textbf{81.42}\textsubscript{0.24} & \color{mgrey}79.56\textsubscript{1.40}\\
BoolQ & \color{mgrey}83.99\textsubscript{3.90} & \color{mgrey}87.14\textsubscript{1.60} & \color{mgrey}87.64\textsubscript{1.10} &  \textbf{88.70}\textsubscript{0.15} & \color{mgrey}\textbf{87.83}\textsubscript{0.10} & \color{mgrey}90.02\textsubscript{0.60} & \textbf{90.15}\textsubscript{0.54} & \color{mgrey}87.77\textsubscript{1.17} & \color{mgrey}64.40\textsubscript{22.37} & \color{mgrey}\textbf{90.10}\textsubscript{0.22}\\
CB & \color{mgrey}45.71\textsubscript{10.61} & \color{mgrey}29.64\textsubscript{7.85} & \color{mgrey}65.71\textsubscript{3.20} & \textbf{81.07}\textsubscript{9.42} & \color{mgrey}\textbf{78.21}\textsubscript{3.19}  & \color{mgrey}\textbf{92.86}\textsubscript{2.19} & \color{mgrey}85.72\textsubscript{7.78} & \color{mgrey}\textbf{92.86}\textsubscript{2.82} &\color{mgrey}89.29\textsubscript{7.25} & \textbf{93.21}\textsubscript{1.49}\\
COPA & \textbf{96.40}\textsubscript{2.30} & \color{mgrey}95.80\textsubscript{2.05} & \textbf{96.40}\textsubscript{2.88} & \color{mgrey}\textbf{96.20}\textsubscript{2.05} & \textbf{96.40}\textsubscript{2.07} & \color{mgrey}\textbf{98.60}\textsubscript{1.14} &  \color{mgrey}97.20\textsubscript{1.10}& \color{mgrey}97.40\textsubscript{0.89} &\textbf{99.00}\textsubscript{0.71}  & \color{mgrey}97.00\textsubscript{1.00}\\
RTE & \color{mgrey}\textbf{80.94}\textsubscript{1.29} & \color{mgrey}79.78\textsubscript{0.92} & \color{mgrey}76.82\textsubscript{1.72} & \color{mgrey}80.43\textsubscript{1.07} & \textbf{83.47}\textsubscript{1.10} & \color{mgrey}75.09\textsubscript{2.11} & \color{mgrey}80.00\textsubscript{2.48} & \color{mgrey}79.21\textsubscript{1.95} & \textbf{86.64}\textsubscript{2.62} & \color{mgrey}\textbf{85.42}\textsubscript{2.48}\\
WiC & \color{mgrey}50.69\textsubscript{0.59} & \color{mgrey}50.56\textsubscript{0.50} & \color{mgrey}49.97\textsubscript{0.13} &\color{mgrey}\textbf{51.38}\textsubscript{3.56}  & \textbf{61.10}\textsubscript{2.07} & \color{mgrey}51.35\textsubscript{1.90} & \color{mgrey}55.58\textsubscript{6.38} & \color{mgrey}54.67\textsubscript{6.02} & \color{mgrey}\textbf{57.87}\textsubscript{11.08} & \textbf{64.83}\textsubscript{8.59} \\
ANLI-R1 & \color{mgrey}\textbf{46.24}\textsubscript{4.21} & \color{mgrey}42.54\textsubscript{3.20} & \color{mgrey}40.26\textsubscript{3.66} &\color{mgrey}40.28\textsubscript{6.46}& \textbf{59.82}\textsubscript{0.51} & \color{mgrey}63.06\textsubscript{2.63} & \color{mgrey}71.92\textsubscript{3.71} & \color{mgrey}\textbf{73.56}\textsubscript{3.88} & \color{mgrey}72.30\textsubscript{8.05} & \textbf{75.00}\textsubscript{3.03}\\
ANLI-R2 & \color{mgrey}40.44\textsubscript{0.90} & \color{mgrey}38.36\textsubscript{0.82} & \color{mgrey}38.44\textsubscript{3.46} & \color{mgrey}\textbf{41.88}\textsubscript{4.50} & \textbf{50.16}\textsubscript{0.82} & \color{mgrey}58.40\textsubscript{1.19} & \color{mgrey}65.36\textsubscript{3.75} & \color{mgrey}\textbf{65.48}\textsubscript{1.91} & \color{mgrey}64.98\textsubscript{2.94} & \textbf{67.30}\textsubscript{2.34}\\
ANLI-R3 & \color{mgrey}42.53\textsubscript{0.99} & \color{mgrey}38.78\textsubscript{1.04} & \color{mgrey}\textbf{43.67}\textsubscript{5.25} & \color{mgrey}37.50\textsubscript{0.81} & \textbf{55.75}\textsubscript{1.66} & \color{mgrey}61.35\textsubscript{3.14} & \textbf{67.32}\textsubscript{0.98} & \color{mgrey}66.23\textsubscript{0.72} & \color{mgrey}63.03\textsubscript{6.03} & \color{mgrey}\textbf{66.38}\textsubscript{0.74}\\
Avg. & \color{mgrey}66.20 & \color{mgrey}62.39 & \color{mgrey}64.98 & \color{mgrey}\textbf{67.65} & \textbf{74.41} & \color{mgrey}75.16 & \color{mgrey}\textbf{77.83} & \color{mgrey}76.89 & \color{mgrey}76.13 & \textbf{81.09}
\end{tblr}}
\end{table}
\begin{figure}[t!]
\vspace{-0.3cm}
    \centering
    \includegraphics[width=\linewidth]{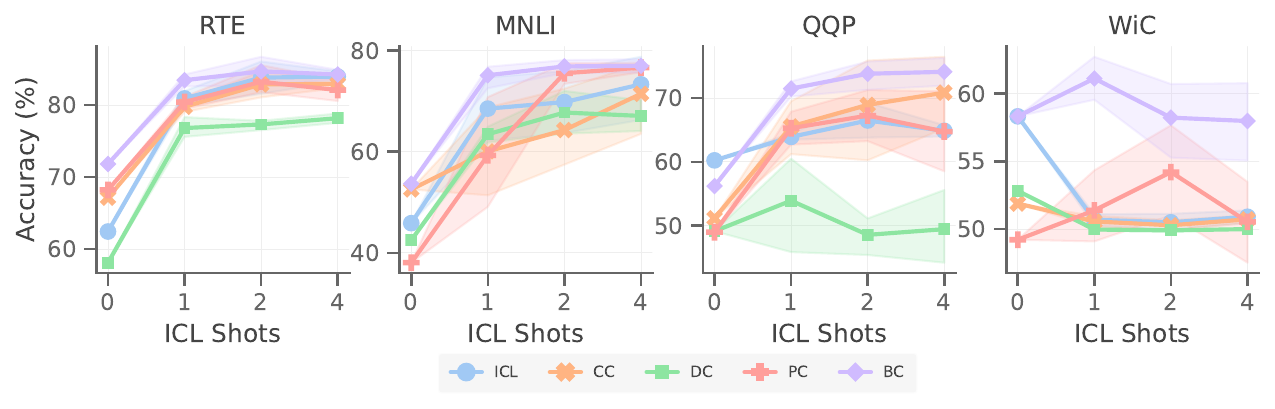}
    \vspace{-0.7cm}
    \caption{The ICL performance on various calibration techniques over the number of ICL shots on PaLM 2-S. Each shot indicates 1 example per class in the demonstration. Lines and shades denote the mean and standard deviation over 5 random seeds, respectively. }
    \label{fig:shots}
\end{figure}
\section{Experiments}
\subsection{Experimental Setup}
\paragraph{Evaluation Data.}
For natural language tasks, in contrast to previous works that only report on relatively simple single-sentence classification tasks \citep{zhao2021calibrate, fei-etal-2023-mitigating, han2023prototypical}, we conduct experiments on 13 more diverse and challenging classification tasks, including the standard GLUE \citep{wang-etal-2018-glue} and SuperGLUE \citep{wang2019superglue} datasets. Specifically, we consider commonsense reasoning: BoolQ \citep{clark-etal-2019-boolq}, COPA \citep{roemmele2011choice}; word disambiguation: WiC \citep{pilehvar-camacho-collados-2019-wic}; sentiment classification: SST-2 \citep{socher-etal-2013-recursive}; paraphrasing: QQP, MRPC \citep{dolan-brockett-2005-automatically}; natural language inference and entailment: ANLI-R\{1,2,3\} \citep{nie-etal-2020-adversarial}, CB \citep{de2019commitmentbank}, RTE, QNLI (QA/NLI), MNLI \citep{williams-etal-2018-broad}.  For image classification tasks, we include SVHN \citep{yuval2011reading}, EuroSAT \citep{helber2019eurosat}, and CLEVR \citep{johnson2017clevr}.

\paragraph{Models.}
We conduct experiments mainly on the state-of-the-art PaLM 2 \citep{anil2023palm} for its variants with different sizes, PaLM 2-S, PaLM 2-M, and PaLM 2-L. PaLM 2 has achieved results that are competitive with GPT-4 \citep{openai2023gpt4}, and readers are referred to \cite{anil2023palm} for more model details. For VLMs, we report the results on CLIP ViT-B/16 \citep{radford2021learning}.
\subsection{Main Experiments}
\begin{table}[t!]
\centering
\caption{Accuracy (\%) on natural language classification tasks with the zero-shot BC and the BCL that involves an optimal strength term learned from a labeled set. The experiments are evaluated with the same in-context example on 1-shot PaLM 2-S. }
\centering
\resizebox{\linewidth}{!}{%
\begin{tblr}{
  width = \linewidth,
  column{2-15} = {c},
  vline{2,15} = {-}{},
  hline{1-2,4} = {-}{},
  hline{1,4} = {-}{0.08em},
  stretch = 0.5
}
Method & SST-2 & MNLI & QNLI & MRPC & QQP & BoolQ & CB & COPA & RTE & WiC & ANLI\textsubscript{R1} & ANLI\textsubscript{R2} & ANLI\textsubscript{R3} & Avg. \\
BC     &   95.4   &     75.0      &  83.5 &  68.6    &   70.3    &  87.9     & 75.0 & 98.0 & \textbf{84.1} & \textbf{63.3} & \textbf{59.8} & \textbf{51.1} & \textbf{53.3} & 74.3 \\
BCL  &  \textbf{96.3}    &     \textbf{75.0}      &   \textbf{83.5}  &   \textbf{74.3}   &    \textbf{72.3}     &   \textbf{88.8}     & \textbf{83.9}   &\textbf{99.0}&82.7&63.2&58.0&49.7&52.2& \textbf{75.3}\\
\end{tblr}}\label{tab:bcl}
\end{table}
\label{sec:mainexp}
\paragraph{Experiments on Natural Language Tasks.}\begin{wrapfigure}{r}{0.48\textwidth}
  \begin{center}
     \vspace{-1.1cm}
     \includegraphics[width=\linewidth]{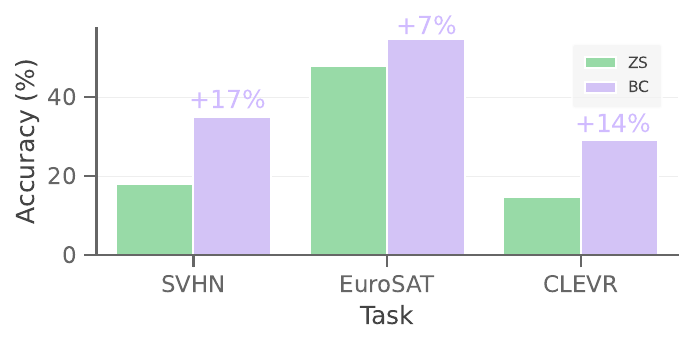}
     \vspace{-0.7cm}
     \caption{\textit{BC improves zero-shot (ZS) image classification: }Accuracy (\%) on image classification tasks with the zero-shot CLIP ViT-16/B. The BC implementation is zero-shot, and we apply BC together with the CLIP to demonstrate the effectiveness of BC in vision-language models. Refer to additional tasks in Appendix~\S\ref{app:aexp}.}\label{fig:clip}
     \vspace{-0.7cm}
\end{center}
\end{wrapfigure}
We present the results across a diverse set of NLP tasks in Table~\ref{tab:main}. Notably, BC consistently outperforms ICL, yielding significant performance enhancement of 8\% and 6\% on PaLM 2-S and PaLM 2-L, respectively. This shows that the BC implementation successfully mitigates the contextual bias from the in-context examples and unleashes the full potential of LLM in efficient learning and quick adaptation to new tasks. In addition, BC improves over the state-of-the-art PC baseline by 6\% on PaLM 2-S, and surpasses the competitive CC baseline by another 3\% on average on PaLM 2-L. Specifically, BC is a generalizable and cheaper technique across all evaluated tasks, delivering stable performance improvement, whereas previous baselines exhibit varying degrees of instability across tasks: DC baseline is the least competitive; CC displays more failure cases in multi-sentence classification tasks, particularly for paraphrasing and NLI tasks, as we hypothesized in Sec~\ref{sec:analysis}; PC, while occasionally competitive, exhibits large performance fluctuations, as evidenced by its large standard deviation, resulting in frequent substantial performance degradation. 

We further analyze the performance of BC by varying the ICL shots from 0 to 4 shots as shown in Fig.~\ref{fig:shots}, and BC again outperforms all baseline methods. We also observe an overall trend for improved performance when more shots are available, and the performance disparities between BC and ICL converge on some tasks, which suggests that BC allows LLMs to more effectively take advantage of more in-context demonstrations. We also observe that PC exhibits the worst stability across the number of shots. In Table \ref{tab:bcl}, we show that further slight gains, 1\% on average, can be achieved by involving an adjustable strength parameter that refines the calibration and mitigates estimation errors. This alternative design not only makes BC applicable to both zero-shot and few-shot setups but also shows its capability to improve further from limited labeled data.

\paragraph{Calibrating Vision-Language Models}
We further extend the applicable scenarios of BC to multi-modal learning to handle the bias inherent in the prompt template designs in CLIP. We select three tasks in which the previous visual-prompting method shows significant improvement \citep{oh2023blackvip}. As shown in Fig.~\ref{fig:clip}, BC significantly improves the zero-shot baseline by 12\% on average. This observation further highlights the presence of contextual bias even within vision-language models, and BC can successfully restore the performance of VLM in image classification tasks, suggesting that BC may serve as a versatile and common technique for mitigating contextual biases across multiple modalities.

\begin{figure}[!t]
\vspace{-0.4cm}
   \begin{minipage}{0.485\textwidth}
     \centering
     \includegraphics[width=\linewidth]{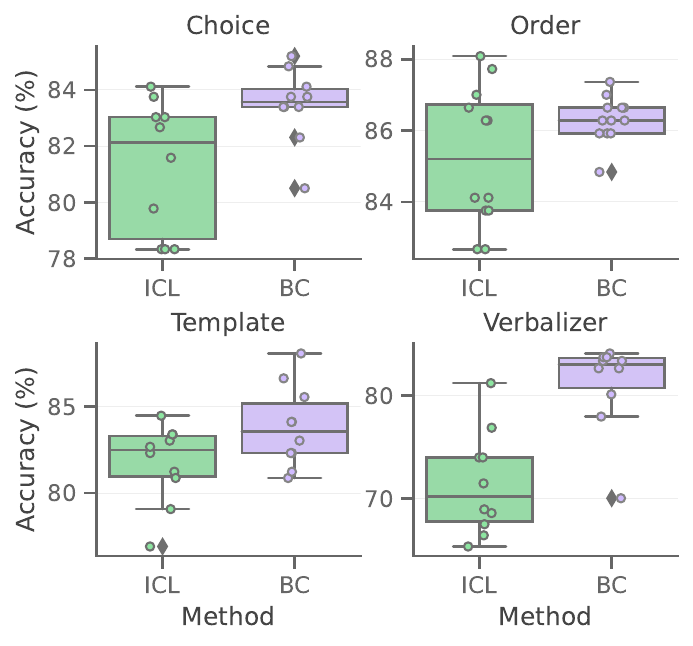}
     \vspace{-0.87cm}
     \caption{\textit{BC makes prompt engineering easier: }Performance of BC with respect to ICL choices, ICL orders, prompt templates, and verbalizers.}\label{fig:robust}
   \end{minipage}\hfill
   \begin{minipage}{0.485\textwidth}
     \centering
     \includegraphics[width=\linewidth]{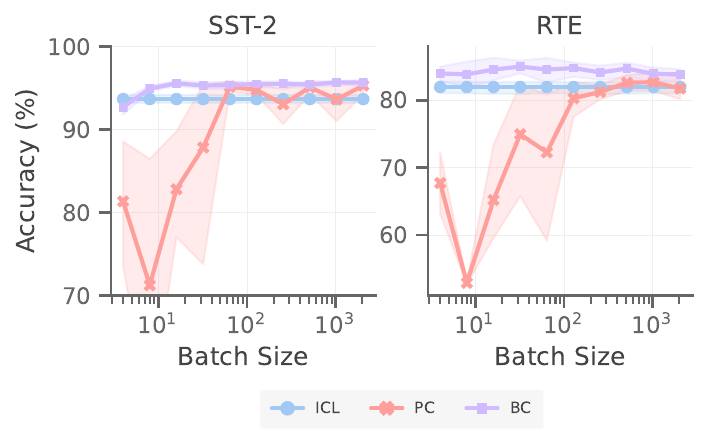}

    \vspace{1.5mm}
     \caption{\textit{BC is data-efficient and insensitive to the batch size: }Performance of BC across different sizes of an initial unlabeled set without using a running estimate of the contextual bias. We compare BC with the state-of-the-art PC baseline that also leverages unlabeled estimate set, and experiments are conducted on PaLM 2-S.}\label{fig:batch}
   \end{minipage}
   \vspace{-0.2cm}
\end{figure}
\subsection{Robustness and Ablation Studies}
\paragraph{Robustness.}
We analyze the robustness of BC with respect to common prompt engineering design choices that were previously shown to significantly affect LLM performance \citep{lu-etal-2022-fantastically, liu-etal-2022-makes}: choices and orders of in-context examples, the prompt template for ICL, and the verbalizers, as shown in Fig. \ref{fig:robust} evaluated on RTE. Setup details are listed in Appendix~\S\ref{app: template}. First, we find that BC is more robust to ICL choices and can mostly achieve the same performance with different ICL examples. Additionally, given a single set of ICL shots, altering the order between each ICL example has minimal impact on the BC performance. However, it is worth noting that an optimal order selection can still lead to promising ICL performance. Furthermore, we analyze the robustness of BC under 10 designs of prompt templates, where BC shows consistent improvement over the ICL baseline. Therefore, though BC makes further improvements, a well-designed template can further enhance the performance of BC. Lastly, we examine the robustness of BC to variations in verbalizer designs. Remarkably, even when employing unconventional choices such as emoji pairs as the verbalizers (listed in Tables~\ref{tab:robustness} \& \ref{tab:verbalizer}) leading to dramatic oscillations of ICL performance, BC largely recovers performance. This observation shows BC robustifies LLM predictions under common prompt design choices and makes prompt engineering easier. 

\paragraph{Batch Size.} We study the impact of batch size on the performance of BC as shown in Fig. \ref{fig:batch}. In contrast to PC, which also leverages an unlabeled estimate set, BC is remarkably more sample efficient, achieving a strong performance with only around 10 unlabeled samples, whereas PC requires more than 500 unlabeled samples before its performance stabilizes.

\section{Conclusion}
We first revisit previous calibration methods while addressing two critical research questions from an interpretation of decision boundaries, revealing their failure cases and deficiencies. We then propose Batch Calibration, a zero-shot and inference-only calibration technique. We also introduce an adjustable extension, BCL, which offers more refined calibrations when labeled data is accessible. While methodologically simple and easy to implement with negligible computation cost, we show that BC scales from a language-only setup to the vision-language context, achieving state-of-the-art performance in both modalities. BC significantly improves the robustness of LLMs with respect to prompt designs, and we expect easy prompt engineering with BC while exploring the potential of BC to generative tasks in the future.

\section*{Acknowledgement}
We thank Emily Salkey for her sincere project management support. We also thank Mohammad Havaei, Chirag Nagpal, Stephen Pfohl, Alexander D'Amour, and Ahmad Beirami for fruitful suggestions and feedbacks and the PaLM 2 team at Google for helping with infrastructure questions. 
\bibliography{iclr2024_conference}
\bibliographystyle{iclr2024_conference}
\clearpage
\appendix
\section{Additional Related Work}
\paragraph{Prompt Learning.}
Prompt learning is an efficient learning pipeline for LLM as an alternative to traditional full-model fine-tuning \citep{liu2023pre}. Soft prompting \citep{li-liang-2021-prefix, lester-etal-2021-power, liu-etal-2022-p} enables fast adaptation of LLM by appending learnable continuous prompts in the embedding space while freezing the rest of weights. The recent development of parameter-efficient fine-tuning methods \citep{houlsby2019parameter, hu2021lora, zhou2023autopeft}, which learn additional modules, may also be interpreted as a form of soft prompting \citep{he2021towards}. However, these soft prompt learning methods inevitably require gradients and internal model access. On the other hand, hard prompting \citep{shin-etal-2020-autoprompt} is an appealing learning category for learning discrete prompts. Recent efforts have been devoted to black-box prompt search without accessing model gradients, and more interpretable prompts can be found by reinforcement learning \citep{deng-etal-2022-rlprompt, zhang2023tempera}, gradient estimation \citep{diao2023blackbox}, and other derivative-free search algorithms \citep{prasad-etal-2023-grips, zhou2023survival}. 
\paragraph{Test-Time Adaptation. } Test-time adaptation aims to mitigate the covariate shift using the test-time statistics. \citet{wang2021tent} propose to minimize the entropy by updating the affine parameters in the BN layer. \citet{nado2020evaluating} and \citet{Schneider2020robust} introduce using test-time batch statistics for the standardization in BN and mixing it with source statistics to conquer covariate shift, respectively. Similarly, mixing the statistics with predefined hyperparameters \citep{you2021test, khurana2021sita}, interpolating source and target-domain statistics \citep{lim2023ttn}, or using a running average estimate \citep{Mirza2022norm, hu2021mixnorm} have also been proposed to adapt the BN layer. \citet{zou2022strength} introduce strength parameters in adapting the standardization statistics in semantic segmentation tasks. We differentiate from test-time BN approaches by mitigating the bias in the novel context of LLM, and there is no source statistic similar to a BN layer in computer vision backbones.

\section{Limitations and Future Work}
BC is a test-time method that relies on the target statistics from batched input. To mitigate any potential estimation errors from unlabelled data, we introduce the adjustable BCL extension to incorporate source statistics from labelled data. Though BC has shown remarkable sample efficiency in terms of batch sizes, it still requires a batch of inputs to estimate the contextual bias. We introduce a running estimation for BC from mini-batches, which subsequently stabilizes the prediction of LLMs when more mini-batches arrive.

In future work, we will endeavor to explore calibration for generative tasks while extending BC to more models across modalities. We suspect that the contextual bias may also exist in short-form generation tasks. Motivated by \citet{zhao2021calibrate}, one possible solution for generative tasks is to calibrate over the logits of the first output token since the following tokens are likely deterministic based on the first token.  Overall, BC is zero-shot, inference-only, and incurs negligible additional costs. We expect easy prompt engineering with BC for users building towards their own robust and responsible LLM applications.

\section{Dataset Statistics}
\begin{table}[H]
\centering
\caption{Details of the dataset used for evaluation in the Table \ref{tab:main}. $|$Test$|$ denotes the number of test samples, where we consistently use the validation split as the test split because labels are not publicly available for some datasets.}
\begin{tblr}{
  cell{2}{3} = {c},
  cell{2}{4} = {c},
  cell{2}{5} = {c},
  cell{3}{3} = {c},
  cell{3}{4} = {c},
  cell{3}{5} = {c},
  cell{4}{3} = {c},
  cell{4}{4} = {c},
  cell{4}{5} = {c},
  cell{5}{3} = {c},
  cell{5}{4} = {c},
  cell{5}{5} = {c},
  cell{6}{3} = {c},
  cell{6}{4} = {c},
  cell{6}{5} = {c},
  cell{7}{3} = {c},
  cell{7}{4} = {c},
  cell{7}{5} = {c},
  cell{8}{3} = {c},
  cell{8}{4} = {c},
  cell{8}{5} = {c},
  cell{9}{3} = {c},
  cell{9}{4} = {c},
  cell{9}{5} = {c},
  cell{10}{3} = {c},
  cell{10}{4} = {c},
  cell{10}{5} = {c},
  cell{11}{3} = {c},
  cell{11}{4} = {c},
  cell{11}{5} = {c},
  cell{12}{3} = {c},
  cell{12}{4} = {c},
  cell{12}{5} = {c},
  cell{13}{3} = {c},
  cell{13}{4} = {c},
  cell{13}{5} = {c},
  cell{14}{3} = {c},
  cell{14}{4} = {c},
  cell{14}{5} = {c},
  hline{1-2,15} = {-}{},
    hline{1,15} = {-}{0.08em},
  stretch=0.5
}
Dataset & Objective               & \#sentences & \#classes & \textbar{}Test\textbar{} \\
SST-2   & Sentence Classification & 1           & 2         & 872                      \\
MNLI    & NLI                     & 2           & 3         & 9815                     \\
QNLI    & Question-Answering NLI  & 2           & 2         & 5463                     \\
MRPC    & Paraphrasing            & 2           & 2         & 408                      \\
QQP     & Paraphrasing            & 2           & 2         & 40430                    \\
BoolQ   & Commonsense Reasoning   & 2           & 2         & 3270                     \\
CB      & NLI                     & 2           & 3         & 56                       \\
COPA    & Commonsense Reasoning   & 3           & 2         & 100                      \\
RTE     & NLI                     & 2           & 2         & 277                      \\
WiC     & Context Comprehension   & 3           & 2         & 638                      \\
ANLI-R1 & NLI                     & 2           & 3         & 1000                     \\
ANLI-R2 & NLI                     & 2           & 3         & 1000                     \\
ANLI-R3 & NLI                     & 2           & 3         & 1200                     
\end{tblr}
\end{table}
\section{Implementation Details}
\paragraph{Contextual Calibration \citep{zhao2021calibrate} (CC).} We follow the original implementation of CC and take the mean of the log-probability over three content-free tokens as the test sample in the predefined template: `N/A', `', `[MASK]'. It incurs 3 additional API costs from LLMs. 
\paragraph{Domain-Context Calibration \citep{fei-etal-2023-mitigating} (DC).} We reproduce the DC baseline by using the same test set as the unlabeled text set to construct its bag-of-words. We then randomly sample tokens for an average length to form the content-free and in-domain input from the bag-of-words. This process is then repeated randomly for 20 times, and we take the mean of the log-probability following the original implementation. It incurs 20 additional API costs from LLMs.
\paragraph{Prototypical Calibration \citep{han2023prototypical} (PC).} For a fair comparison, we use the same test set as the unlabeled estimate set for PC. We follow the same hyper-parameters reported by PC with 100 maximum iterations for EM and 100 times random initialization for the whole learning process to stabilize its estimation. It is noteworthy that this number of repetitions is costly and relatively slow, especially when the $|$Test$|$ is large. 
\paragraph{Batch Calibration (BC).} In all reported experiments, we compute the correction log-probability term $\hat{\mathbf{p}}(y|C)$ once after all test samples are seen. In the $n$-shot ICL experiments reported in Table \ref{tab:main} and Fig. \ref{fig:shots}, the $k$-shot ICL is concatenating $k$ random training sample per class. In the BCL experiment that uses labeled samples, we use $J\times128$ randomly selected training samples as the labeled data. In the robustness study, we use 1 randomly sampled example as the context to study the performance of BC with respect to the ICL choices. We then conduct the ICL order experiment by re-ordering 4 randomly sampled ICL examples. The rest experiments are conducted on the standard 1-shot ICL setup.
\label{appendix:details}
\section{Additional Experiments}
\label{app:aexp}
\begin{table}[!ht]
\centering
\caption{Accuracy (\%) on natural language classification tasks with 0-shot PaLM 2-S and 1-shot PaLM 2-M models in a single seed.}
\label{tab:app_main1}
\resizebox{\linewidth}{!}{%
\begin{tblr}{
  width = \linewidth,
  cell{1}{2} = {c=5}{c},
  cell{1}{7} = {c=5}{c},
  cell{2}{2-11} = {c=1},
  vline{2-3} = {1}{},
  vline{2,7} = {1-17}{},
  hline{1-3,16-17} = {-}{},
    hline{1,17} = {-}{0.08em},
  stretch = 0.5
}
{Model\\ \hfill} & {PaLM 2-S\\0-shot} &  &  &  &  & {PaLM 2-M\\1-shot} &  &  &  & \\
Method & ICL & CC & DC & PC & BC & ICL & CC & DC & PC & BC \\
SST-2 & 94.61 & 94.50 & 94.61 & 87.84 & \textbf{95.18} & 94.95 & 95.87 & 94.95 & \textbf{96.22} & 96.10\\
MNLI & 45.87 & 52.54 & 42.50 & 38.04 & \textbf{53.67} & 45.50 & 54.43 & 56.26 & 43.81& \textbf{60.02}\\
QNLI & 49.28 & 48.97 & 49.44 & \textbf{50.28} & 49.55 & 78.88 & 75.56 & 62.95 & 77.39 & \textbf{78.91}\\
MRPC & 69.12 & 61.76 & \textbf{69.85} & \textbf{69.85} & 64.95 & 57.11 & \textbf{73.53} & 68.87& 69.85& 65.93\\
QQP & \textbf{60.23} & 51.16 & 49.12 & 48.98 & 56.20 & 66.18 & \textbf{79.67} & 74.32& 70.27& 75.13\\
BoolQ & 86.51 & \textbf{86.97} & 76.88 & 55.41 & 84.04 & 87.37 & 88.53 & 87.28& \textbf{88.78}& 87.31\\
CB & \textbf{85.71} & 58.93 & 55.36 & 46.43 & 67.86 & 71.43 & 69.64 & 67.86& 50.00& \textbf{80.36}\\
COPA & 88.00 & 66.00 & \textbf{90.00} & 52.00 & 88.00 & \textbf{97.00} & 96.00 & 96.00& \textbf{97.00}& 96.00\\
RTE & 62.45 & 67.15 & 58.12 & 68.23 & \textbf{71.84} & 77.62 & 79.06 & 68.23& 77.98& \textbf{80.51}\\
WiC & 58.31 & 51.88 & 52.82 & 49.22 & \textbf{58.30} & 61.13 & 64.11 & 52.04& 65.52& \textbf{68.03}\\
ANLI-R1 & 39.80 & 44.70 & 43.00 & 37.00 & \textbf{50.00} & 52.40 & 52.40 & 52.70 & 35.70& \textbf{54.00}\\
ANLI-R2 & 36.80 & 41.50 & 40.70 & 40.20 & \textbf{45.10} & 46.00 & \textbf{50.70} & 47.80 & 35.80 & 50.00\\
ANLI-R3 & 42.67 & 46.42 & 43.08 & 35.50 & \textbf{48.50} & 43.50 & 45.67 & 49.33 & 32.42 & \textbf{50.50}\\
Avg. & 63.03 & 59.42 & 58.88 & 52.23 & \textbf{64.09} & 67.62 & 71.17 & 67.58 & 64.67 & \textbf{72.52}
\end{tblr}}
\vspace{5mm}
\end{table}

\begin{table}[!ht]
\centering
\caption{Accuracy (\%) on image classification tasks with the zero-shot CLIP ViT-16/B. We additionally report on UCF101 \citep{soomro2012ucf101}, FGVC Aircraft \citep{maji2013fine}, and DTD \citep{cimpoi2014dtd}. }
\centering
\begin{tblr}{
  width = \linewidth,
  column{2-8} = {c},
  vline{2,8} = {-}{},
  hline{1-2,4} = {-}{},
  hline{1,4} = {-}{0.08em},
  stretch = 0.5
}
Method & SVHN & EuroSAT & UCF & CLEVR & Aircraft & DTD & Avg. \\
ZS     &   18.0   &     47.8      &   \textbf{66.7}  &  14.7    &    \textbf{24.8}     &   \textbf{44.4}     & 36.7   \\
ZS+BC  &  \textbf{35.0}    &     \textbf{54.7}      &   66.0  &   \textbf{29.2}   &    22.3     &   41.7     & \textbf{41.5}    \\
\end{tblr}\label{tab:clip}
\vspace{-1.2cm}
\end{table}

\begin{table}[!ht]
\vspace{15mm}
\centering
\caption{Performance of BC with respect to prompt templates and verbalizers per example in Fig.~\ref{fig:robust}. We report accuracy (\%) on the RTE task on 1-shot PaLM 2-S. We refer each example ID to the index of templates and verbalizers listed in Tables \ref{tab:template} \& \ref{tab:verbalizer}.}
\begin{tblr}{
  width = \linewidth,
  column{3} = {c},
  column{4} = {c},
  column{5} = {c},
  column{6} = {c},
  column{7} = {c},
  column{8} = {c},
  column{9} = {c},
  column{10} = {c},
  column{11} = {c},
  column{12} = {c},
  cell{1}{1} = {c=2}{},
  cell{2}{1} = {r=2}{},
  cell{4}{1} = {r=2}{},
  cell{2}{1} = {r=2}{},
  cell{4}{1} = {r=2}{},
  vline{2} = {1}{},
  vline{2-3} = {2-5}{},
  vline{3} = {3,5}{},
  hline{1-2,4,6} = {-}{},
  hline{1,6} = {-}{0.08em},
  stretch = 0.5
}
ID &  & 1 & 2 & 3 & 4 & 5 & 6 & 7 & 8 & 9 & 10\\
Template & ICL & 81.2 & 80.9 & 82.3 & 76.9 & 83.0 & 83.4 & 83.4 & 82.7 & 79.1 & 84.5\\
 & BC & 84.1 & 80.9 & 82.3 & 85.6 & 84.1 & 81.2 & 88.1 & 82.3 & 83.0 & 86.6\\
Verbalizer & ICL & 81.2 & 76.9 & 74.0 & 74.0 & 67.5 & 69.0 & 71.5 & 68.6 & 65.3 & 66.4\\
 & BC & 84.1 & 83.4 & 83.4 & 80.1 & 78.0 & 70.0 & 82.7 & 82.7 & 83.8 & 83.8
\end{tblr}
\label{tab:robustness}
\end{table}

\begin{figure*}[t!]
  \vspace{-0.3cm}
  \begin{subfigure}{}
    \centering\includegraphics[width=0.95\linewidth]{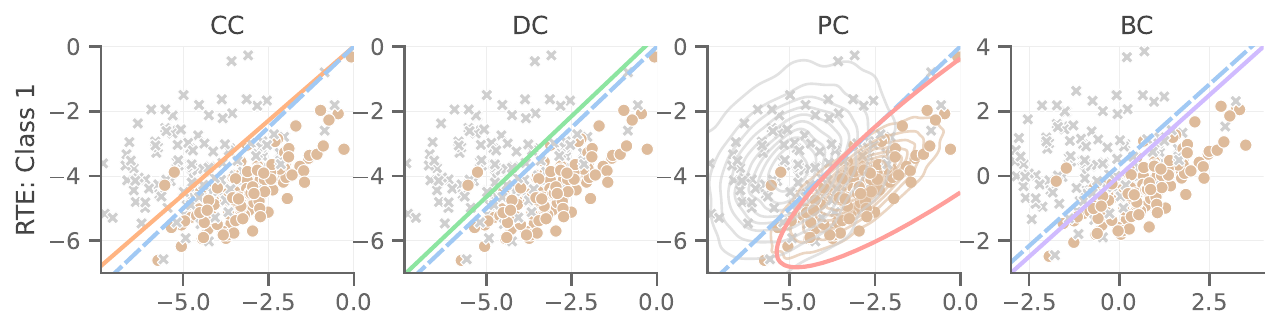}
    \caption*{}
  \end{subfigure}
  \vspace{-1.2cm}
  \begin{subfigure}{}
    \centering\includegraphics[width=0.95\linewidth]{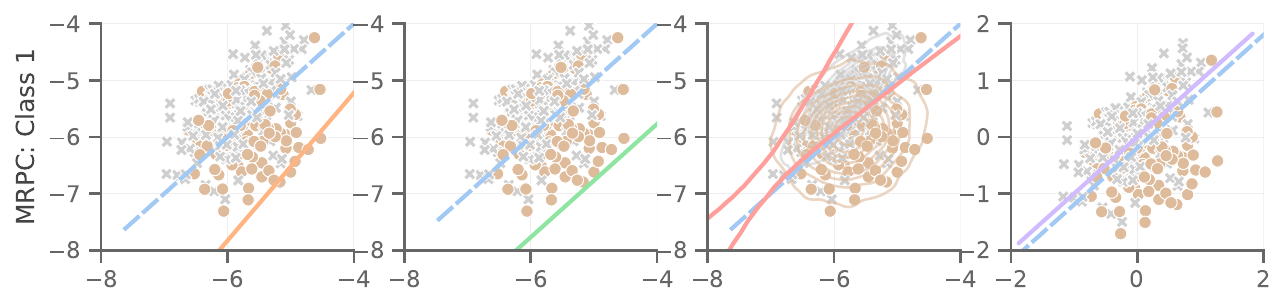}
    \caption*{}
  \end{subfigure}
    \vspace{-1.2cm}
  \begin{subfigure}{}
    \centering\includegraphics[width=0.95\linewidth]{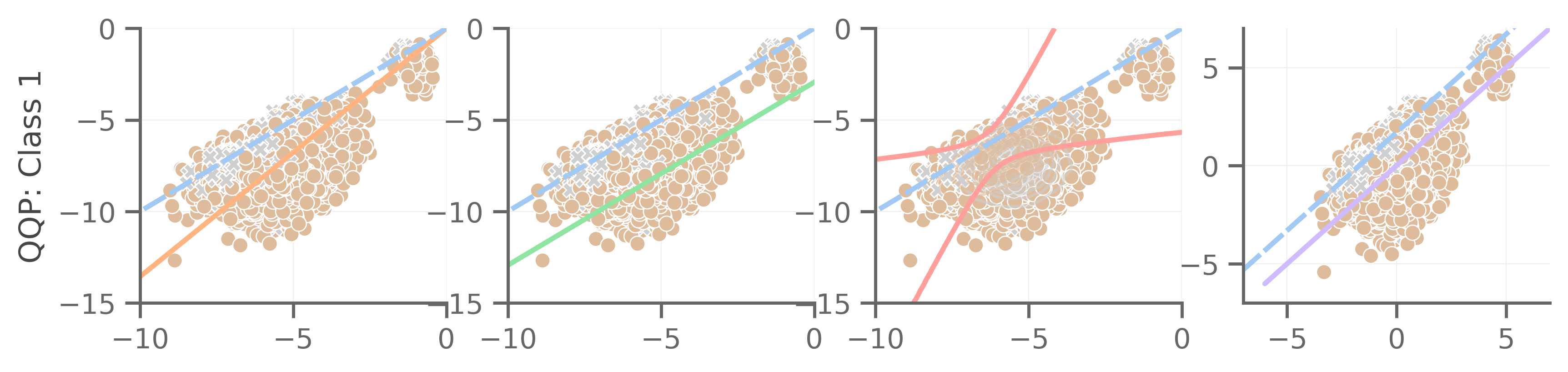}
    \caption*{}
  \end{subfigure}
    \vspace{-1.2cm}
  \begin{subfigure}{}
    \centering\includegraphics[width=0.95\linewidth]{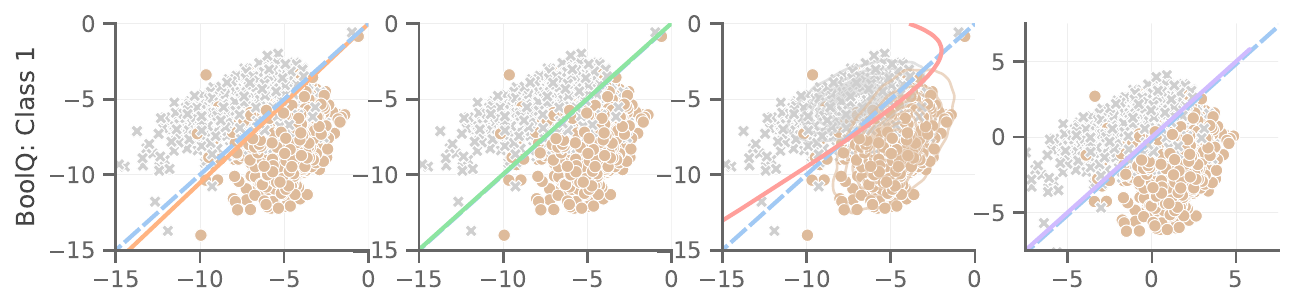}
    \caption*{}
  \end{subfigure}
    \vspace{-1.2cm}
  \begin{subfigure}{}
    \centering\includegraphics[width=0.95\linewidth]{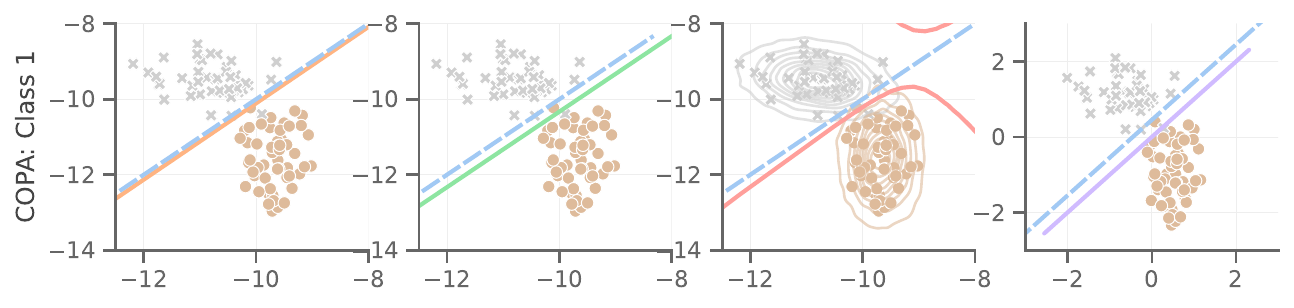}
    \caption*{}
  \end{subfigure}
    \vspace{-1.2cm}
  \begin{subfigure}{}
    \centering\includegraphics[width=0.95\linewidth]{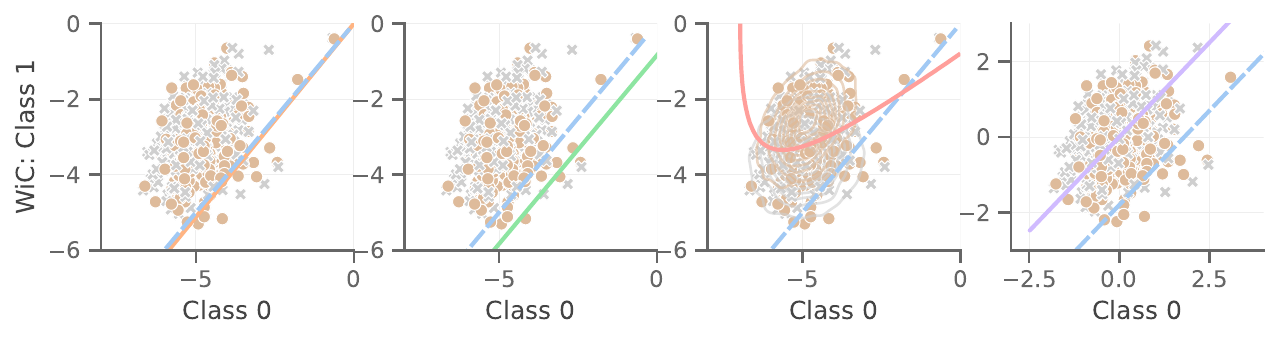}
    \caption*{}
  \end{subfigure}
  \vspace{-1cm}
  \caption{Visualization of the decision boundaries of uncalibrated ICL, and after applying existing calibration methods and the proposed BC. We list all binary classification tasks from the evaluation set.}
  \label{fig:dba}
\end{figure*}

\begin{figure*}
     \centering
     \vspace{-0.9cm}
     \includegraphics[width=\linewidth]{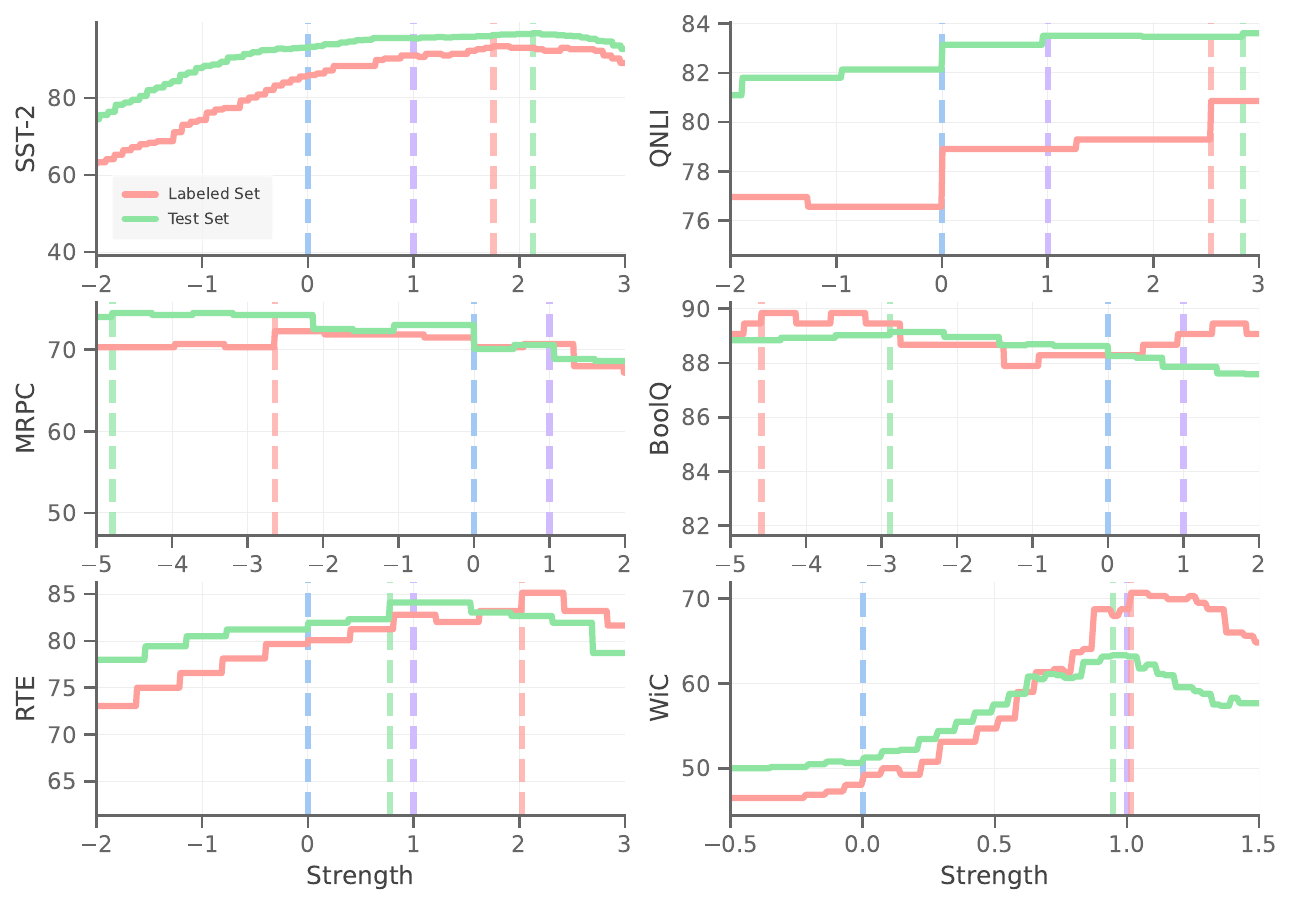}
     \caption{The performance of an adaptable batch calibration layer (BCL) compared to the zero-shot BC with a changing strength. The \texttt{strength} $\gamma$ at 0 and 1 represent the uncalibrated ICL and BC, respectively. We highlight the optimal strength learned from a labeled set by a red vertical line and the best test strength by a green line.}\label{fig:bcl2}
\end{figure*}

\clearpage
\section{Prompt Templates}
\label{app: template}
\begin{table}[h]
\centering
\caption{Prompt templates for all $k$-shot ICL experiments. We follow the template styles from \citet{han2023prototypical} and \citet{brown2020language}.}
\begin{tblr}{
  width = 0.8\linewidth,
  colspec = {Q[83]Q[688]Q[167]},
  hline{1-11} = {-}{},
  hline{3-10} = {-}{0.03em},
  hline{1,11} = {-}{0.08em},
}
Dataset          & Template                                                                                & Label Set           \\
SST-2            & {Review: \{sentence\}\\Sentiment: \{label\}}                                            & {negative / \\positive} \\
{MNLI\\CB\\ANLI} & {Premise: \{premise\}\\Hypothesis: \{hypothesis\}\\Answer: \{label\}}                   & yes / maybe / no    \\
QNLI             & {Question: \{question\}\\Sentence: \{sentence\}\\Label: \{label\}}                      & yes / no            \\
MRPC             & {Sentence 1: \{sentence1\}\\Sentence 2: \{sentence2\}\\Equivalence: \{label\}}          & no / yes            \\
QQP              & {Question 1: \{question1\}\\Question 2: \{question2\}\\Duplicate: \{label\}}            & no / yes            \\
BoolQ            & {\{passage\}\\Question: \{question\}\\Answer: \{label\}}                                & no / yes            \\
COPA             & {Premise: \{premise\}\\Choice1: \{choice1\}\\Choice2: \{choice2\}\\Answer: \{label\}}   & 1 / 2               \\
RTE              & {Premise: \{sentence1\}\\Hypothesis: \{sentence2\}\\Answer: \{label\}}                  & yes / no            \\
WiC              & {Sentence1: \{sentence1\}\\Sentence2: \{sentence2\}\\Word: \{word\}\\Answer: \{label\}} & false / true        
\end{tblr}
\end{table}

\begin{table}[ht!]
\centering
\caption{Prompt templates for the robustness experiment conducted on RTE in Fig. \ref{fig:robust}. }
\begin{tblr}{
  width = \linewidth,
  colspec = {Q[40]Q[787]Q[112]},
  cell{2}{3} = {r=10}{},
  vline{2-3} = {-}{},
  hline{1-2,12} = {-}{},
  hline{3-11} = {1-2}{},
  hline{3-11} = {-}{0.03em},
  hline{1,12} = {-}{0.08em},
  stretch=0.5
}
ID & Template                                                                                                                                              & Label Set \\
1  & {Premise: \{sentence1\}\\Hypothesis: \{sentence2\}\\Answer: \{label\}}                                                                                & yes / no  \\
2  & {\{sentence1\}\\Hypothesis: \{sentence2\}\\Answer: \{label\}}                                                                                         &           \\
3  & {\{sentence1\}\\Question: \{sentence2\}\\Answer: \{label\}}                                                                                           &           \\
4  & {\{sentence1\}\\Question: \{sentence2\}\\\{label\}}                                                                                                   &           \\
5  & {\{sentence1\}\\Question: \{sentence2\}\\yes or no? Answer: \{label\}}                                                                                &           \\
6  & {Sentence 1: \{sentence1\}\\Sentence 2: \{sentence2\}\\Answer: \{label\}}                                                                             &           \\
7  & {Premise: \{sentence1\}\\Hypothesis: \{sentence2\}\\Label: \{label\}}                                                                                 &           \\
8  & {Sentence 1: \{sentence1\}\\Sentence 2: \{sentence2\}\\Label: \{label\}}                                                                              &           \\
9  & {Determine if the sentence 2 is true based on the Sentence 1 below\\Sentence 1:~\{sentence1\}\\Sentence 2:~\{sentence2\}\\Answer:~\{label\}}          &           \\
10 & {Determine if the sentence 2 is true or false based on the Sentence 1 below\\Sentence 1:~\{sentence1\}\\Sentence 2:~\{sentence2\}\\Answer: \{label\}} &           
\end{tblr}
\label{tab:template}
\end{table}

\begin{table}[ht!]
\centering
\caption{Verbalizer choices for the robustness experiment conducted on RTE in Fig. \ref{fig:robust}, where we include emoji pairs for ID 8, 9, 10. }
\begin{tblr}{
  width = \linewidth,
  colspec = {Q[62]Q[419]Q[437]},
  cell{2}{3} = {r=10}{},
  vline{2-3} = {-}{},
  hline{1-2,12} = {-}{},
  hline{3-11} = {1-2}{},
  hline{3-11} = {-}{0.03em},
  hline{1,12} = {-}{0.08em},
  stretch=0.5
}
ID & Label Set                  & Template                                                               \\
1  & yes / no                   & {Premise: \{sentence1\}\\Hypothesis: \{sentence2\}\\Answer: \{label\}} \\
2  & true / false               &                                                                        \\
3  & correct / incorrect        &                                                                        \\
4  & positive / negative        &                                                                        \\
5  & good / bad                 &                                                                        \\
6  & great / terrible           &                                                                        \\
7  & it was true / it was false &                                                                        \\
8  & :thumbs\_up / :thumbs\_down    &                                                                        \\
9  & :man\_gesturing\_ok / :man\_gesturing\_no  &                                                                        \\
10 & :check\_mark / :cross\_mark   &
\end{tblr}
\label{tab:verbalizer}
\end{table}

\begin{table}[h]
\centering
\caption{Prompt templates for the 0-shot experiments.}
\begin{tblr}{
  width = \linewidth,
  colspec = {Q[83]Q[688]Q[167]},
  hline{1-2,11} = {-}{},
  hline{3-10} = {-}{0.03em},
  hline{1,11} = {-}{0.08em},
}
Dataset          & Template                                                                                                                             & Label Set           \\
SST-2            & {Review: \{sentence\}\\Sentiment: \{label\}}                                                                                         & {negative / \\positive} \\
{MNLI\\CB\\ANLI} & {\{premise\}\\Question: \{hypothesis\} yes, no, or maybe?\\Answer: \{label\}}                                                        & yes / maybe / no    \\
QNLI             & {\{question\}\\Question: \{sentence\} yes or no?\\Answer: \{label\}}                                                                 & yes / no            \\
MRPC             & {Sentence 1: \{sentence1\}\\Sentence 2: \{sentence2\}\\Equivalence: \{label\}}                                                       & no / yes            \\
QQP              & {Question 1: \{question1\}\\Question 2: \{question2\}\\Duplicate: \{label\}}                                                         & no / yes            \\
BoolQ            & {\{passage\}\\Question: \{question\}\\Answer: \{label\}}                                                                             & no / yes            \\
COPA             & {Premise: \{premise\}\\Choice1: \{choice1\}\\Choice2: \{choice2\}\\Answer: \{label\}}                                                & 1 / 2               \\
RTE              & {\{sentence1\}\\Question: \{sentence2\} yes or no?\\Answer: \{label\}}                                                               & yes / no            \\
WiC              & {\{sentence1\}\\\{sentence2\}\\Question: Is the word '\{word\}' used in the same way in the two sentences above?\\Answer: \{label\}} & no / yes            
\end{tblr}
\end{table}
\end{document}